# Satellite downlink scheduling under breakpoint resume mode

Zhongxiang Chang [1,2,※], Yuning Chen[3], Zhongbao Zhou[1,2]

[1]School of Business Administration, Hunan University, Changsha, China, 410082

[2]Hunan Key Laboratory of intelligent decision-making technology for emergency management, Changsha, China, 410082

[3]School of System Engineering, National University of Defense Technology, Changsha, China, 410073

## Abstract

A novel problem called satellite downlink scheduling problem (SDSP) under breakpoint resume mode (SDSP-BRM) is studied in our paper. Compared to the traditional SDSP where an imaging data has to be completely downloaded at one time, SDSP-BRM allows the data of an imaging data be broken into a number of pieces which can be downloaded in different playback windows. By analyzing the characteristics of SDSP-BRM, we first propose a mixed integer programming model for its formulation and then prove the NP-hardness of SDSP-BRM. To solve the problem, we design a simple and effective heuristic algorithm (SEHA) where a number of problem-tailored move operators are proposed for local searching. Numerical results on a set of well-designed scenarios demonstrate the efficiency of the proposed algorithm in comparison to the general purpose CPLEX solver. We conduct additional experiments to shed light on the impact of the segmental strategy on the overall performance of the proposed SEHA.

## 1. Introduction

The earth observation satellite (EOS) plays an important role in environmental monitoring, land surveys and detailed investigation of sensitive areas and other fields. Satellite mission planning and scheduling problem mainly contains two parts: data acquisition task scheduling and data downlink scheduling. The data acquisition is an imaging activity, while the data downlink is a playback activity as shown in Figure 1.

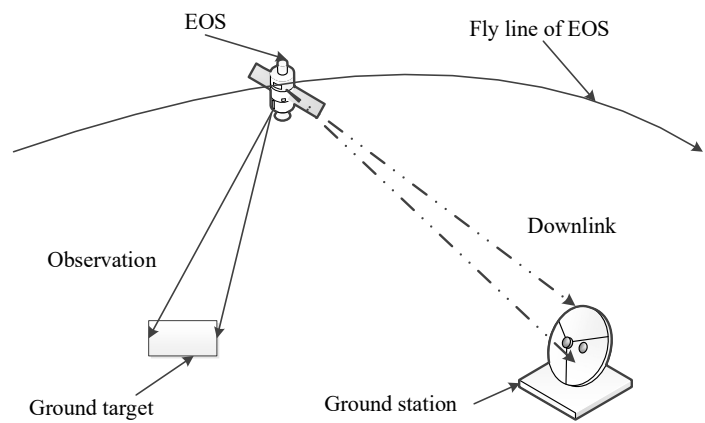

Figure 1 Observation and Downlink

When the data acquisition activity is completed, the corresponding imaging data will be stored in the satellite, then which need to be transmitted to the receiving resources (like ground stations) by data downlink activities. In theory, the data of each downlink activity could be a partial (incomplete) imaging data, a complete imaging data or even a combination of multiple complete or partial imaging data.

With the development of space technology, the imaging capability of satellites has been greatly enhancing, which causes a big explosion in the amount of imaging data. GAOFEN II (Huang et al., 2018), launched in 2014, marks the arrival of "submeter Era" for EOS in China. While the downlink capacity of satellite antennas does not develop synchronously. The downlink rate of GAOFEN II antenna is only 2×450Mbps. There is a big gap, the imaging data obtained by one-second observation will spend 4.5 seconds to play back, between the data acquisition capability and the data download capability. The big gap poses a new challenge to SDSP.

SDSP and its variations have been studied by many authors. Some of these works were focused on a single satellite (Karapetyan et al., 2015, Peng et al., 2017, Chang et al., 2020, Song et al., 2019b) whereas others were more general purpose in the special satellite constellation or multi-satellite (Bianchessi et al., 2007, Wang et al., 2011, Al et al., 2004). On the other hand, some researchers saw SDSP as a time-dependent or resource-dependent problem and focused on the time-switch constraints between satellites and ground stations (Du et al., 2019, Marinelli et al., 2011, Verfaillie et al., 2010, Zhang et al., 2019, Zufferey et al., 2008, Chang et al., 2022, Lu et al., 2021, Chang et al., 2021b, Chang et al., 2021a, Chang et al., 2020). In addition, other authors (Hao et al., 2016, Li et al., 2014, Peng et al., 2017, Wang et al., 2011, Zhao et al., 2019, Cho et al., 2018) saw image data as an uncertain process, they considered satellite observation planning (SOP) and SDSP together or transformed SDSP as constraints for SOP. Chen (Xiao-yue, 2009) abstracted the possible positions of digital tasks in a scheduling sequence as nodes, and constructed a matrix solution construction graph of the pheromone distribution in the nodes. To solve it they proposed an ant colony algorithm. Chang (CHANG Fei 2010) considered SDSP as a complex constrained combinatorial optimization problem, and proposed a particle swarm optimization algorithm with controllable speed, direction and size. Chen (Chen et al., 2016) regarded the data acquisition chain of satellite mission planning as a path planning problem with multi peak features, and established a framework for solving the problem based on the label a constrained shortest path method. Giuseppe (Corrao et al., 2012) proposed a method to solve multiple satellites and multiple stations planning for automatically arranging tasks from the perspective of ground station. Chen (Hao et al., 2016) built a cooperative scheduling model considering observation and transmission for electronic reconnaissance satellites, and proposed a method based on the genetic algorithm for solving it. Li (LI Yun-feng, 2008) established the descriptive model of satellite data transmission tasks and the satellite data transmission scheduling model. To solve the problem, they designed a hybrid genetic algorithm. Chen (Chen et al., 2015) established a satellite data transmission scheduling model which is suitable for practical application, and a quantum discrete particle swarm optimization algorithm was proposed to solve the problem. Maillard (Maillard et al., 2016) presented a data transmission scheduling method for SDSP based on cooperation between satellite and ground in the presence of uncertain imaging data, and made contrast experiments with complete ground SDSP and Pure Onboard SDSP, results of which showed the method based on corporation had significant advantages. Li (Li et al., 2014) regarded SDSP as a multi-constraint, multi-object and multi-satellite data transmission scheduling problem, and established a data transmission scheduling topology model. They solved the problem with the K-shortest path genetic algorithm.

Through a limited review about the researches according to SDSP, we find most of them are concerned about the allocation of playback window resources and resolving conflicts between playback windows. Especially, there is a similar assumption in their studies that one playback window can transmit multiple images and for a single image the data cannot be separated (Karapetyan et al., 2015) and have to be transmitted in "First observed, First downlink (FOFD)" order (Wang et al., 2011). Under these conditions, SDSP is equivalent to the satellite range scheduling problem (SRSP) (Luo et al., 2017, Marinelli et al., 2011, She et al., 2019, Song et al., 2019a, Chu and Chen, 2018). But because of the above instruction that there is a big gap between the data acquisition capability and the data download capability, the assumption is no longer practical. In this paper, we make the first attempt to address the problem called satellite data downlink scheduling problem under breakpoint resume mode (SDSP-BRM). The breakpoint resume mode allows the transmission of an image data pause at some point and resume later on. Under this mode, the data of a single image can be divided into a number of small pieces. The purpose of the SDSP-BRM is to arrange the whole images or their small pieces into playback windows in order to maximize the total reward. Because of this additional dimension of complexity, SDSP-BRM is no longer a simple downlink request permutation problem (DRPP) (Karapetyan et al., 2015). SDSP-BRM is more complicated than SRSP (Zufferey et al., 2008, Vazquez and Erwin, 2014, Barbulescu et al., 2004) and SDSP (Wang et al., 2011, She et al., 2019), both of which are NP-Hard (Barbulescu et al., 2004, Vazquez and Erwin, 2014), so SDSP-BRM is NP-Hard, too.

The rest of this paper is organized as follows. In section 2, we will present a mathematical formulation of SDSP-BRM and provide an analysis of the problem complexity. In section 3, we will present a simple and effective heuristic algorithm (SEHA) for solving SDSP-BRM. Then experimental results and analysis are reported in section 4 and concluding remarks are given in section 5.

## 2. Problem Analysis

In this section, we present the inputs and outputs, constraints and assumptions of SDSP-BRM firstly. Then we explain the relationship between SDSP-BRM and the Knapsack problem to prove the SDSP-BRM is a NP-Hard problem. Finally, a mixed integer programming model is proposed to formulate SDSP-BRM.

## 2.1. The input and output of the SDSP

In the following, we give formal definitions of the inputs, image data and playback windows, and outputs, playback tasks, of SDSP(-BRM).

### 2.1.1. Imaging data

Imaging data ($t$) refers to the data acquired by the sensors of EOS, which can be represented by a six-element tuple below:

$$t = \langle n, p, os, oe, od, d \rangle \tag{1}$$

where $n$ denotes the identity number of $t$. $p$ reflects the priority of $t$. $os$, $oe$ and $od$ represents start time, end time and image duration of the image data $t$ respectively. $d$ denotes the duration for downloading $t$.

Since imaging data obtained by one-second observation will spend 4.5 seconds to play back as mentioned in section 1, the relationship between $od$ and $d$ is:

$$d = 4.5 \times od \tag{2}$$

Therefore, the six-element tuple description can be simplified as a five-element tuple: $t = \langle n, p, os, oe, d \rangle$.

### 2.1.2. Playback window

The playback window ($w$) refers to the visible time window between the satellite and receiving resources (like the ground stations and the relay satellites), which can be represented by a five-element tuple:

$$w = \langle m, f, ds, de, l \rangle \tag{3}$$

where $m$ denotes the identity number of $w$. $f$ indicates the type of the receiving resources. $f = 0$ represents $w$ is playback window of a ground station, while $f = 0$ represents the receiving resources is a relay satellite. $ds$, $de$ and $l$ reflects start time, end time and duration of $w$ respectively.

With the development of the relay satellites, the bandwidth of which develops from the Ku band to Ka band (Li et al., 2014), so the data receiving capabilities of the ground station and the relay satellite are roughly the same, therefore, the description of the data playback window can be simplified as a four-element tuple, $w = \langle m, ds, de, l \rangle$.

### 2.1.3. Playback task

As mentioned above, the playback task ($TS$) is the output of SDSP-BRM, which is the action of satellites to playback the imaging data, and can be represented by a four-element tuple below:

$$TS = \langle m, ts, te, set \rangle \tag{4}$$

where $m$ denotes the identity number of $TS$, which is directly inherited from the playback window. $ts$ and $te$ indicates the execution start time and end time of $TS$ respectively. And $set$ represents a set of the identity number of all imaging data transmitted in the playback window $m$.

### 2.2. Assumptions

(1) Satellite has a file system to management all imaging data (Huang et al., 2018), so we assume that the imaging data can be transmitted discontinuously.

(2) If only if an imaging data can be transmitted completely, there is possible to select it for scheduling. Otherwise the imaging data should be abandoned directly.

(3) There is no any setup time for playback tasks.

### 2.3. Constraints

### 2.3.1. Service constraint

A playback window can service an imaging data means that the window can be used to transmit the imaging data. We use service coefficient ($r_{ij} = 0$ or $1$) expressing service of each window with each imaging data, and there is a simple algorithm for calculating service coefficient following:

```
For i =1: Number of Imaging data
    For j=1: Number of Windows
        If ds_j > oe_i  then
            r_{i,j} = 1 ;
        Else
            r_{i,j} = 0 ;
        End if
    End For
End For
```

As shown in of Figure 2, we can obtain the service coefficient according to the above-mentioned algorithm: $r_{11}=1$、$r_{21}=0$、$r_{31}=0$、$r_{12}=1$、$r_{22}=1$、$r_{32}=0$.

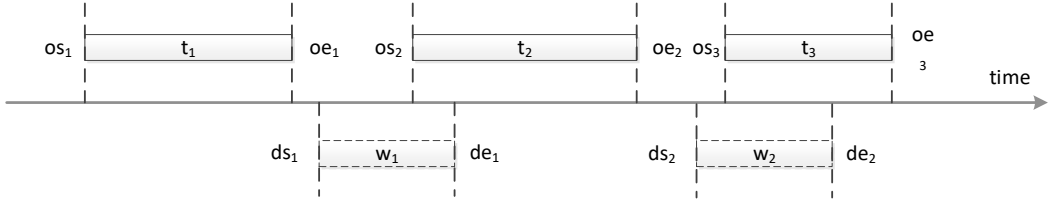

Figure 2 Demonstration of service constraint

### 2.3.2. Imaging data segmentable constraint

The imaging data is stored in the on-board storage system as a continuous memory unit. In theory, the imaging data can be divided into several infinite pieces. But for the practical application, the infinite segmentation is unacceptable, therefore the minimum length ($ld$) of imaging data segmentable is proposed. When the minimum length is added, the segment process can be expressed as the following logic:

```
For i =1: Number of Imaging data
    While (other constraints are met)
        If Segment(t_i) < ld  then
            Break;// Cannot be segmented
        Else
            Segment(t_i);
        End if
    End while
End For
```

where $\text{Segment}(t_i)$ represents the process of segmenting the imaging data.

### 2.4. Problem Complexity Analysis

If the imaging data can be segmented, SDSP becomes more challenging. Figure 3 illustrates the difference between SDSP with segmental strategy or not. Figure 3 (a) shows that when the data download capability is comparable to the data acquisition capability. The focus of SDSP is how to schedule various imaging data and obtain more data under the limited playback windows, which is similar to the classic Knapsack problem that has been proved to be a NP-Hard problem (Garey and Johnson, 1979).

However, with the synchronous development of the data acquisition capability and the data downlink capability, one playback window cannot transmit an imaging data completely.

Therefore, considering the segmental strategy is imperative. Figure 3 (b) shows SDSP considering segmental strategy, which includes two processes: segmenting the imaging data and allocating the playback windows to transmit the segmented imaging data.

SDSP-BRM is a completely new problem, which can decompose into three sub-problems:

(1) Decide whether an imaging data is transmitted or not;

(2) Decide how to segment the imaging data;

(3) Decide how to use limited playback windows to transmit all segmented imaging data.

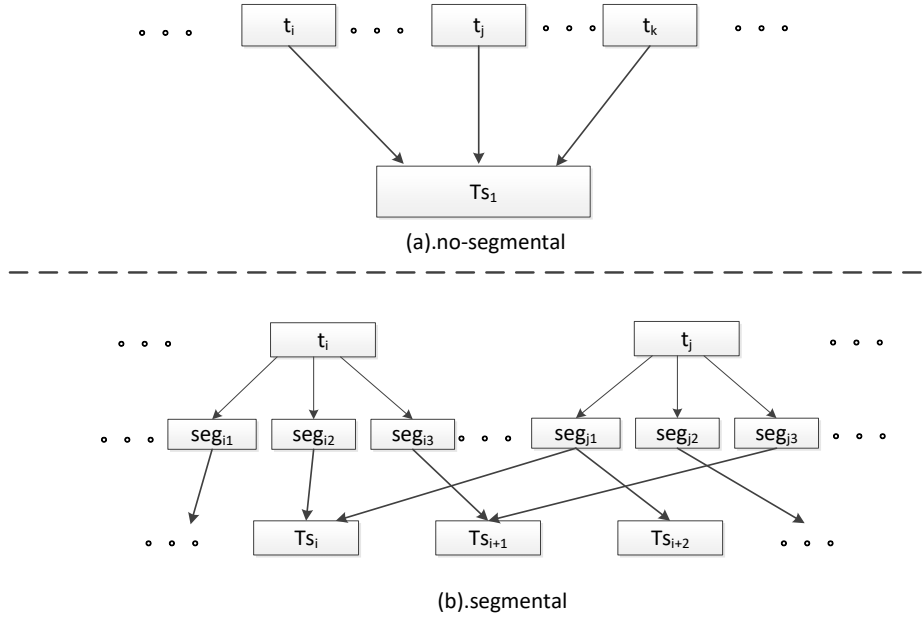

Figure 3 SDSP with segmental strategy or not

In the field of operations research, there are many classical problems like those of sub-problems. The first sub-problem is a typical 0-1 integer programming problem; the second sub-problem is similar to the Cutting Stock problem (CSP), which is a NP-hard proved in theory (Delorme et al., 2016); the third sub-problem is similar to the Bin Packing problem (BPP) or the Knapsack problem, which has the typical characteristics of integer optimization. Moreover, they are closely related and interacted with each other. Therefore, the SDSP-BRM is more complex and is a NP-Hard problem.

**2.5. Mixed integer programming model for SDSP-BRM**

The optimal object of SDSP is to arrange a sequence of downlink tasks to maximize the number of imaging data transmitted under the condition of receiving resources shortage and many constraints (Karapetyan et al., 2015).

Let $T = \{t_i | i = 1,2,\cdots,N\}$ denotes a collection of imaging data needs to be transmitted, wherein N represents the total number of imaging data and $t_i = \langle n_i, p_i, os_i, oe_i, d_i \rangle$.

Let $W = \{w_j | j = 1,2,\cdots,M\}$ indicates the set of the playback windows, wherein M represents the total number of playback windows and $w_j = \langle m_j, ds_j, de_j, l_j \rangle$.

As described above, SDSP-BRM can be divided into three sub problems. Therefore, the model has following four decision variables.

$x_i$ indicates whether $t_i$ is selected for transmit, which is a 0-1 integer variable. In addition, this parameter corresponds to the sub-problem one;

$$x_i = \begin{cases} 1 & if\ t_i\ is\ selected \\ 0 & otherwise. \end{cases}$$

$y_{i,j}$ is a non-negative real number, which is a continuous variable and represents the amount of imaging data of $t_i$ transmitted by $w_j$. Note that, $y_{i,j}$ belongs to the interval [0,1]. In addition, this parameter corresponds to the sub-problem two;

$q_j$ indicates whether $w_j$ is used, which is a 0-1 integer variable;

$$q_j = \begin{cases} 1 & if\ w_j\ is\ used \\ 0 & otherwise. \end{cases}$$

$g_{i,j}$ indicates whether using $w_j$ to transmit $t_i$, which is also a 0-1 integer variable. In addition, the parameters $q_j$ and $g_{i,j}$ correspond to the sub-problem three.

$$g_{i,j} = \begin{cases} 1 & if\ using\ w_j\ to\ transmit\ t_i \\ 0 & otherwise. \end{cases}$$

To transmit as much image data as possible is the original intention of SDSP. Several optimization objective functions, maximize transmission revenue (Karapetyan et al., 2015, Marinelli et al., 2011), maximize downlink duration (Zhang et al., 2019) and minimize transmission failure rate (Du et al., 2019), were adopted in many researches. Without loss of generality, in this paper, we use maximizing transmission revenue as our optimal objective. Therefore, the model is represented as follows:

$$\text{Maximize } \sum_{i=1}^{N} x_i \times p_i \quad (5)$$

S.T.

$$g_{i,j} \times ld \leq y_{i,j} \leq g_{i,j} \times d_i \quad (6)$$

$$\sum_{j=1}^{M} y_{i,j} = x_i \times d_i \quad (7)$$

$$\sum_{i=1}^{N} y_{i,j} \leq l_j \quad (8)$$

$$g_{i,j} \leq r_{i,j} \quad (9)$$

$$g_{i,j} \leq x_i \quad (10)$$

$$y_{i,j} \geq 0 \quad (11)$$

$$g_{i,j} = 0 \text{ or } 1 \quad (12)$$

$$x_i = 0 \text{ or } 1 \quad (13)$$

The right side of the constraint equation (6) represents the upper bound of segmentable pieces for each imaging data, and the left represents the lower bound. a) $g_{i,j} = 1$ donates using $w_j$ to transmit $t_i$, and the equation (6) deform to $ld \leq y_{i,j} \leq d_i$. b) $g_{i,j} = 0$ means $t_i$ does not play data playback at $w_j$, and the equation deform to $y_{i,j} = 0$.

The constraint equation (7) corresponds to the second assumption. $x_i = 1$ means the sum of the sub imaging data must be equal to the imaging data $t_i$, and the equation is deformed to $\sum_{j=1}^{M} y_{i,j} = d_i$. While $x_i = 0$ expresses that $t_i$ is not selected to be transmitted and the equation deform to $\sum_{j=1}^{M} y_{i,j} = 0$.

The constraint equation (8) describes an objective constraint of each playback window that the sum of playback time used must less than or equal to length of the window.

The constraint equation (9) describes the relationship between the service coefficient($r_{i,j}$) and the decision variable ($g_{i,j}$) that $r_{i,j} = 1$ is a necessary condition for $g_{i,j} = 1$.

The constraint equation (10) similar to the constraint equation (9) describes the relationship between the decision variable($x_i$) and the decision variable($g_{i,j}$) that $x_i = 1$ is a necessary condition for $g_{i,j} = 1, j = 1 \dots M$

The constraint equation (11) means that $y_{i,j}$ is a non-negative continuous variable.

The constraint equations (12) and (13) indicate $x_i$ and $g_{i,j}$ is a 0-1 integer variable.

## 3. A Simple and Effective Heuristic Algorithm (SEHA) For SDSP-BRM

As mentioned above, SDSP-BRM is a NP-hard problem, so there is not any exact algorithm can achieve the optimal solution in polynomial time generally. To deal with the large size practical problems, we design a simple and effective heuristic algorithm (SEHA) based on some heuristic rules according to the characteristics of SDSP-BRM and the pseudo-code of the algorithm as following:

> Initialize parameters;
> Construct a feasible solution based on a heuristic greedy algorithm;
> While (Both *Max_Iter and Solve_Time* are not met) do
>     While (*NoUp_Iter* is not met) do
>         Apply the *remove* operator to constructed solution;
>         Apply the *insert* operator to constructed solution;
>         If (The value of the objective function increase)
>             Update the solution;
>             The time number of *no improve* reset;
>         Else
>             The time number of *no improve* add one;
>         End if
>     End while
> End while

Then we want to explain how to construct an feasible solution, how to improve the solution and how to terminate the search algorithm.

### 3.1. Solution construction

Two heuristic rules are designed according to the characteristic of the problem: prefer to select the imaging data with a greater contribution rate and prefer to use the playback window with a smaller service coefficient.

#### 3.1.1. Rule 1 (Prefer to select the imaging data with a greater contribution rate)

The contribution rate($c_i$) of an imaging data($t_i$) is defined as the equation (14), which considers both the priority and the playback time of $t_i$. The rule 1 represents the higher value of $c_i$ is, the higher probability to select $t_i$ is.

$$c_i = \frac{p_i/l_i}{\max_{t_j \in T} p_j/l_j} \quad t_i \in T \tag{14}$$

#### 3.1.2. Rule 2 (Prefer to use the playback window with a smaller service coefficient)

The service coefficient ($r_{.j}$) of a playback window ($w_j$) is defined as the equation (15). The rule 2 means that the smaller value of $r_{.j}$ is, the greater probability to adopt $w_j$ is.

$$r_{.j} = \sum_{t_i \in T} r_{ij} \tag{15}$$

Based on the two heuristic rules above, a heuristic greedy algorithm is designed to construct feasible solutions fast, the basic flow of which is as shown below:

> Step 1 Judge whether the imaging data set T and the playback window set W is empty. If $T = \emptyset$ or $W = \emptyset$, the algorithm terminates, otherwise, go to step 2;

Step 2 Calculate $c_i$ of each $t_i$ and $r_j$ of each $w_j$, then sort in increasing order of $t_j$ over $c_i$ and sort in decreasing order of $w_j$ over $r_j$;

Step 3 Select an imaging data from the series of $t_j$ sorted, then arrange playback windows to transmit $t_j$ from head of the series of $w_j$ satisfying the constraints of (6)-(13);

Step 4 The algorithm terminates when all imaging data are examined.

### 3.2. Move operators

The local search algorithm improves quality of a solution by changing the structure of the solution by a series of move operators. We design two important move operators for SEHA: remove operator and insert operator, as shown in Figure 4.

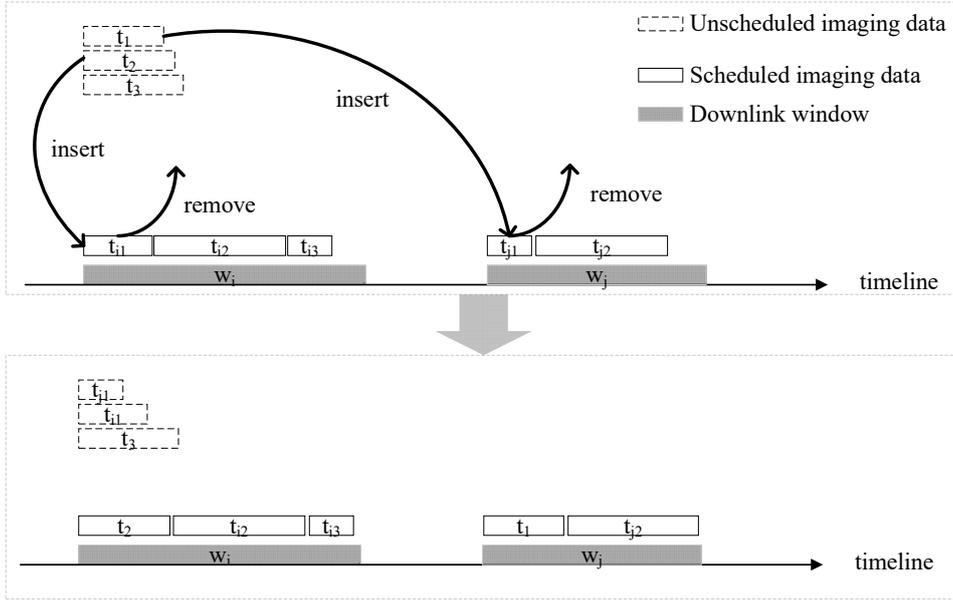

Figure 4 Two types of operator

Use the remove operators to delete some scheduled imaging data from a given scheme randomly, which will produce more unused space in some data playback windows, and then use insert operator inserts to select some unscheduled imaging data randomly for adding in the scheme. After the process of move operators, the structure of the scheme will be changed and a new solution will be generated. If the value of the object improves, the new solution will be accepted.

### 3.3. Termination criterions

We design three termination criterions for SEHA: 1) if the iterations of SEHA reach Max_Iter, SEHA will be terminated; 2) if the solution has no improvement after NoUp_Iter iterations, SEHA will be terminated; 3) if the run time of the algorithm reach Solve_Time, SEHA will be terminated.

## 4. Experimental Study

In this section we design three simulation experiments to explain the efficiency of SEHA. Firstly, we will compare SEHA and CPLEX based on many different scale simulation scenarios. The second experiment illustrates the necessity for considering the proposed heuristic rules and explains the impact of the initial solution to SEHA. In the end, we demonstrate the performance impact of the segmental strategy.

### 4.1. Experimental setup

There is no any benchmark about SDSP in the literature. Therefore, considering on the actual project (GAOFEN II) and the universality of experiment, we design the several different scale scenarios randomly. Table 1 shows their principle to generate every element of the input data, the structure of which are proposed in the section above.

Table 1 Generation principle of scenario

| Input data | Variable | Generation principle |
|---|---|---|
| $t_i$ | $p_i$ | $p_i \sim U[1,10]$ |
| | $os_i$ | $os_i = oe_{i-1} + N(100,1)$ |
| | $d_i$ | $d_i \sim U[2ld, 10ld]$ |
| | $oe_i$ | $oe_i = os_i + d_i/4.5$ |
| $w_j$ | $ds_j$ | $ds_j = de_{j-1} + N(100,1)$ |
| | $l_j$ | $l_j \sim U[ld, 5ld]$ |
| | $de_j$ | $de_j = ds_j + l_j$ |

Since the lack of the data receiving resources, we set the relationship between the number of data window (M) and the number of imaging data (N) as N=a*M, wherein $a \sim U[1.5, 2.5]$. Using the above generation principle, we design several different size scenarios randomly.

SEHA is coded in C++ and the version of CPLEX is 12.5. And both of them run on a PC with Intel i7-3520M (2.90GHz) CPU and 12.0 GB RAM under Windows 7.

### 4.2. Comparative results of SEHA with CPLEX

After analyzing extensive experiments, we obtain the setting rule of parameters of SEHA (*Max_Iter*, *NoUp_Iter* and *Solve_Time*) shown in the Table 2.

Table 2 Parameters of SEHA

| Max_Iter | NoUp_Iter | Solve_Time(s) |
|---|---|---|
| 100000 | 5000 | 60 |

The comparative results of different scale simulation scenarios are shown in Table 3. Among them, *SN* denotes the serial number of each scenario. *N* indicates the number of imaging data in each scenario. *M* reflects the number of data playback window in each scenario. $R_{max}$, $R_{min}$ and $\bar{R}$ denotes the best, worst and mean profit obtained by SEHA and CPLEX respectively. $\bar{T}$ indicates the mean running time of SEHA and CPLEX. $Gap_R$ represents the difference between the best profit ($R_{max}$) of CPLEX and SEHA. If $Gap_R$ is a negative number, $R_{max}$ of SEHA is better than that of CPLEX, and vice versa. And $Gap_{\bar{R}}$ denotes the difference between the mean profit ($\bar{R}$) of CPLEX and SEHA. If $Gap_{\bar{R}}$ is a negative number, $\bar{R}$ of SEHA is better than that of CPLEX, and vice versa. From the experimental results in the Table 3, we can find some interesting phenomena.

Table 3 Comparative results of scenarios

| SN | N | M | CPLEX | | | | SEHA | | | | $Gap_R$ | $Gap_{\bar{R}}$ |
|---|---|---|---|---|---|---|---|---|---|---|---|---|
| | | | $R_{max}$ | $R_{min}$ | $\overline{R_{MIP}}$ | $\bar{T}$ | $R_{max}$ | $R_{min}$ | $\overline{R_{LS}}$ | $\bar{T}$ | | |
| 1 | 20 | 8 | **74** | **74** | **74** | 2 | **74** | 74 | 74 | **0.07** | **0** | 0 |
| 2 | 30 | 15 | **86** | **86** | **86** | 4 | **86** | 84 | 85.21 | **0.22** | **0** | 0.79 |
| 3 | 50 | 24 | 148 | 148 | 148 | 60 | 146 | 142 | 144.47 | **0.69** | 2 | 1.53 |
| 4 | 100 | 70 | 286 | 284 | 285 | 60 | 282 | 277 | 279.19 | **3.5** | 4 | 5.81 |
| 5 | 200 | 85 | 443 | 443 | 443 | 60 | 452 | 444 | 447.23 | **15.57** | **-9** | **-14.23** |
| 6 | 500 | 220 | 564 | 564 | 564 | 60 | 1332 | 1319 | 1322.15 | 60 | **-768** | **-758.15** |
| 7 | 800 | 340 | -- | -- | -- | 60 | 2105 | 2091 | 2096.78 | 60 | -- | -- |
| 8 | 1000 | 530 | -- | -- | -- | 60 | 2903 | 2885 | 2891.49 | 60 | -- | -- |

- SEHA can obtain the optimal solution (i.e., scenario 1 and 2), and the running time (*Solve_Time*) of SEHA is significantly less than that of CPLEX.
- The profit of solution obtained by SEHA is approximate to that obtained by CPLEX (i.e., scenario 3 and 4), while the running time (*Solve_Time*) of SEHA is constantly better than that of CPLEX.
- Note that, much better solutions can be obtained by SEHA than that of CPLEX in acceptable running time (*Solve_Time*=60 s) with the scale of task, like scenario 5 and scenario 6.
- With the size of task increases (i.e., scenario 7 and 8), especially, CPLEX already cannot obtain a feasible solution in acceptable running time (*Solve_Time*), while SEHA can still find satisfactory solutions or even near-optimal solutions.

- The value of $R_{max}$, $R_{min}$ and $\bar{R}$ obtained by SEHA are very close for each scenario, which represents the robustness of SEHA is good.

Given all that above, comparing with CPLEX, SEHA can always obtain good quality solutions (satisfactory solutions or even near-optimal solutions) with fewer running time for different scale simulation scenarios for SDSP-BRM, especially with the scale of task increases.

### 4.3. The influence of heuristic rules

In this section, we carry out two compared experiments to illustrate the influence of the heuristic rules. In the first experiment, we compare the quality of construction solutions using heuristic rules and not using them. Then using these construction solutions in the first experiment as the initial solutions for SEHA, and comparing the final results.

### 4.3.1. The influence of the heuristic rule on the quality of construction solutions

In order to illustrate the efficiency of two proposed heuristic rules, we set up four simulation experiments based on eight examples designed in section 4.2. These four simulation experiments are considered as: Considering both of two heuristic rules (shortly denoted by a & b), only considering the heuristic Rule 1 (shortly denoted by a &!b), only considering the heuristic Rule 2 (shortly denoted by !a & b) and without considering both of those two rules (shortly denoted by !a&!b).

As shown in Figure 5, we can find that the construction solutions considering heuristic rules solution are always much better than those not considering the heuristic rules, and the first rule is more effective than the second one. Notably, as the scale of problem increases, the influence of the heuristic rules is more obvious.

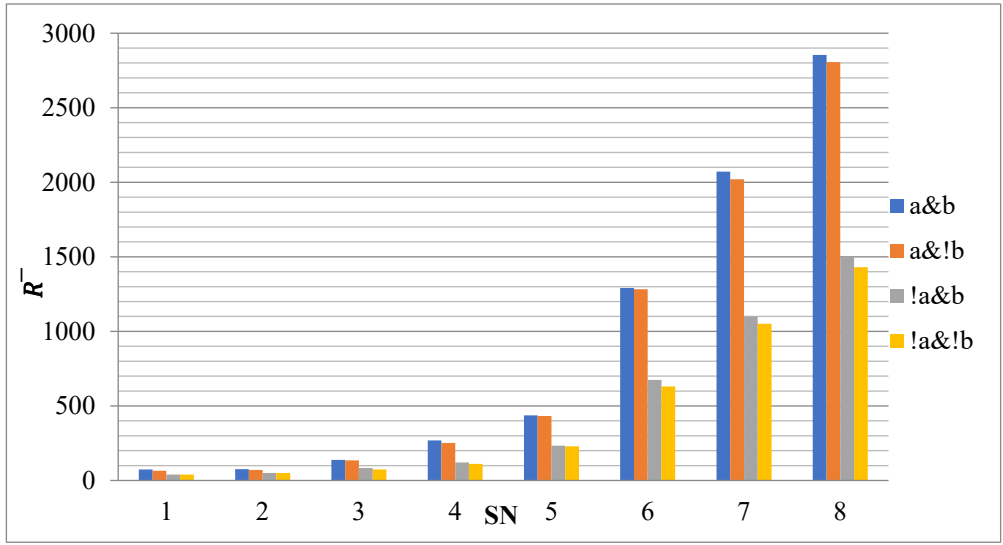

Figure 5 Comparison results of the heuristic Rules

### 4.3.2. The influence of initial solution on SEHA

It is a consensus that the local search algorithm is sensitive to the initial solution (Wu et al., 2013). In this section we will explain the reasonableness of using the construction solution considering the heuristic rules as the initial solution of SEHA.

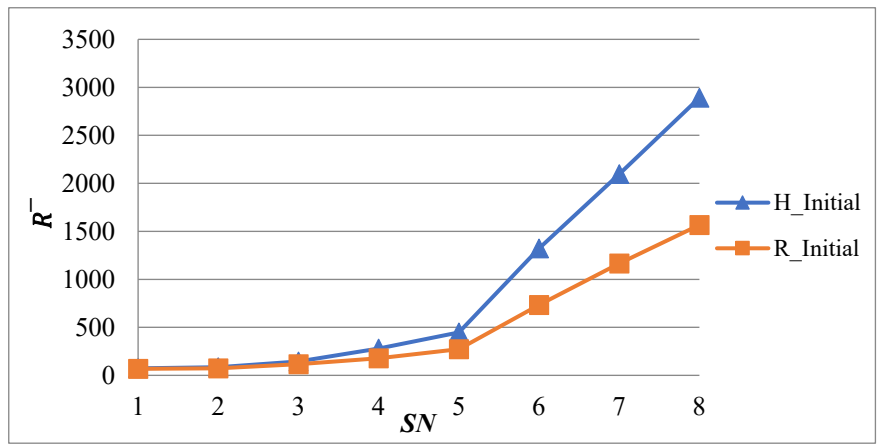

Figure 6 Comparison result of different initial solution

Through the above experiment we know that the construction solution considering the heuristic rules solution is always better. In this experiment we compare the results of SEHA based on the different initial solutions, one of which is based on the heuristic rules (shortly denoted by H_Initial) and the other of which is a randomly generated solution (shortly denoted by R_Initial). As shown in Figure 6, we can find that there is assuredly a strong correlation between the quality of results obtained by SEHA and the quality of initial solutions. So, using construction solution based on the heuristic rules as the initial solution is necessary and reasonable.

### 4.4. The impact of the segmental strategy

To evaluate the performance improvement brought by the segmental strategy, we perform contrast experiments and analysis results obtained by SEHA considering the segmental strategy and not considering it (shortly denoted by SG and Non-SG).

As shown in Figure 7, the segmental strategy significantly improves the profit of satellite playback. And the impact of segmental strategy is more obvious with the scale of problem increases. There are two reasons to explain the phenomenon. 1) Considering the segmental strategy ensures that some imaging data, which has a large contribution rate but cannot be transmitted by any single playback window, can be transmitted. 2) Some playback windows might have some residual space after scheduling without considering the segmental strategy, while considering the strategy we can divide some imaging data to fill those playback windows. Therefore, it is necessary for SDSP to consider the segmental strategy under the condition that the data acquisition capability and data download capability develops asynchronously.

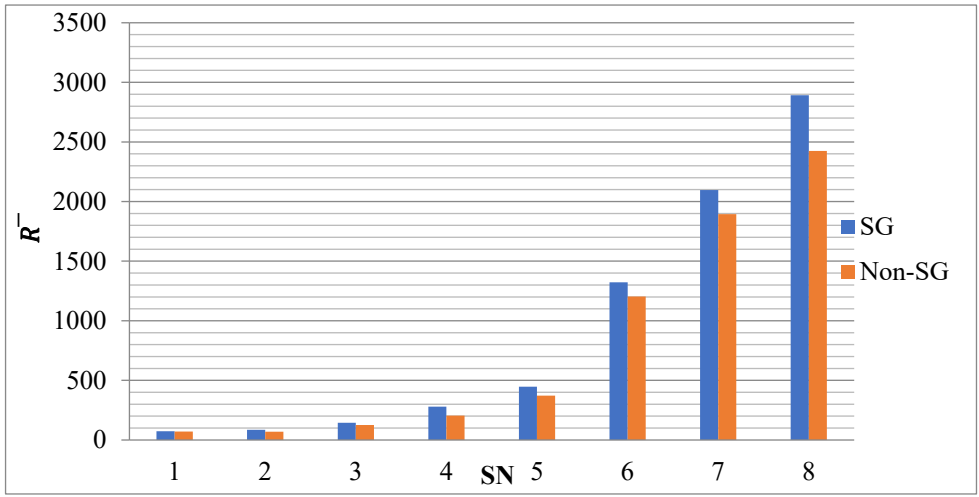

Figure 7 The impact of the segmental strategy

## 5. Conclusions and future work

Based on the analysis of the problem, we prove SDSP-BRM is a NP-Hard problem and establish a mixed integer programming model to formulate it, then we design a simple and effective heuristic algorithm (SEHA) to solve it. Extensive experiments demonstrate that SEHA can find optimal or near-optimal solutions in the acceptable running time (Solve_Time) for different scale simulation instances. While these two heuristic rules will improve the efficiency of SEHA apparently, and the segmental strategy can improve the profit of satellite downlink, and more significantly with the scale of problem increases.

The future work in our study is three-fold: 1) Considering more actual practical demands from engineering, add the time constraint of imaging data transmitted into the model. 2) In this paper, we only consider the mode of data playback, considering the various working modes is an in-depth research direction. 3) SDSP-BRM in this paper is an off-line scheduling, in another word, the imaging data acquisition schedule is given and not changed before downlink scheduling. However, with the development of satellite autonomy, the online scheduling considering the uncertainty of the imaging data acquisition schedule is an interesting direction.

## Acknowledgements

The research of Zhongxiang Chang was supported by the science and technology innovation Program of Hunan Province (2021RC2048)